# DWT Based Fingerprint Recognition using Non Minutiae Features

**Shashi Kumar D R[1], K B Raja[2], R K Chhotaray[3], Sabyasachi Pattanaik[4]**

[1]**Department of CSE, Cambridge Institute of Technology,**
**Bangalore, Karnataka, India**

[2]**Department of ECE, University Visvesvaraya College of Engineering, Bangalore University,**
**Bangalore, karnataka, India**

[3]**Department of CSE, Seemanta Engineering College,**
**Mayurbhanj, Orissa, India**

[4]**Department of Computer Science, F.M. University,**
**Balasore, Orissa, India**

**Abstract**
Forensic applications like criminal investigations, terrorist identification and National security issues require a strong fingerprint data base and efficient identification system. In this paper we propose DWT based Fingerprint Recognition using Non Minutiae (DWTFR) algorithm. Fingerprint image is decomposed into multi resolution sub bands of LL, LH, HL and HH by applying 3 level DWT. The Dominant local orientation angle $\theta$ and Coherence are computed on LL band only. The Centre Area Features and Edge Parameters are determined on each DWT level by considering all four sub bands. The comparison of test fingerprint with database fingerprint is decided based on the Euclidean Distance of all the features. It is observed that the values of FAR, FRR and TSR are improved compared to the existing algorithm.

## 1. Introduction

Identifying an individual based on certain physiological or behavioral characteristics is Biometry. It is most widely used and developed in forensic and non forensic applications in national and corporate security. Fingerprint, palm print, iris, cornea, face, voice print, gait and DNA are some of the biological features of persons used in biometrics.

Fingerprints are developed at fetal stage and do not change throughout one's life. The identification system based on fingerprint recognition can substitute the unconventional system based on pass word, certificate or personal Identification number. Fingerprint is a patterned impression consisting of ridges and valleys. A ridge is a single curved segment and valley is the region between two adjacent ridges. Uniqueness of the fingerprint depends upon the pattern of ridges and the minutiae point locations. Minutiae points are local ridge characteristics that occur either at ridge bifurcation or at ridge ending. Fingerprints are normally classified into five major classes viz., arch, tented arch, left loop, right loop and whorl.

Among all the biometric traits, fingerprint identification technology is well developed. The uniqueness, immutability and low cost of fingerprint system have made it to be most widely used and being universally accepted as the best identification system. The widespread application of fingerprint was in government agencies and law enforcement cells till recently. However the availability of inexpensive fingerprint capturing devices, high speed computing hardware, explosive growth of internet transactions and efficient identification algorithms has created huge market for the fingerprint authentication system which is used for personal identification in almost every security applications.

The fingerprint system has two types, identification and verification. Identification is one to many comparison of test fingerprint with database to identify the unknown person whereas verification is one to one comparison of a test fingerprint with stored template to verify the claimed person. Usually Identification take more processing time as the test fingerprint is to be compared with all the fingerprint templates of the data base. Fingerprint authentication systems have broadly four approaches, minutiae based approach [1], Ridge based verification [2], pore Extraction [3] and image based approach [4]. Minutia based approach demands very expensive processing techniques like binarization, noise removal, normalization, segmentation, orientation estimation, ridge filtering, thinning etc,. and it is time consuming. Ridge based approach requires extraction of ridges in the fingerprint and finding the distance between them which will be normally tedious. The level three fingerprint feature like pores is very distinctive and there are two types of pores viz., closed and open pores. Closed pore is enclosed by a ridge and an open pore intersects with the valley lying between the two ridges. But image based approach requires minimum preprocessing time since the minutiae, ridges and pores are not considered.

*Contribution:* In this paper, the 3 level DWT is applied on fingerprint images. The Directional Information features like Coherence and Dominant local orientation angle θ, Centre Area features and Canny's Edge parameters are computed from DWT sub bands. The Euclidean Distance is used to verify the test Fingerprint with data base fingerprint.

*Organization:* The paper is organized into the following sections. Section 2 is an overview of related work. The DWTFR model is described in Section 3. Section 4 is the algorithm for DWTFR system. Performance analysis of the system is presented in Section 5 and Conclusions are contained in Section 6.

## 2. Related Work

Honglie Wei and Danni liu [5] have proposed a fingerprint matching technique based on three stages of matching which includes local orientation structure matching, local minutiae structure matching and global structure matching. The similarity of neighborhood minutiae distribution around minutiae is found by local minutiae structure matching. The global similarity is evaluated by global minutiae structure. The final similarity is found by the matching score calculated from the local orientation structure matching, local minutiae structure and global structure matching. Chengming wen et al. [6] has proposed an algorithm for one-to-one matching of minutiae points using motion coherence methods. The K-plet was used to describe local structure.

Ujjal Kumar Bhowmik et al. [7] proposed that smallest minimum sum of closest Euclidean distance (SMSCED) corresponding to the rotation angle to reduce the effect of non linear distortion. The overall minutiae patterns of the two fingerprints are compared by the SMSCED between two minutiae sets. Khuram Yasin Qureshi and Shoab A.Khan [8] proposed fingerprint matching using five neighbor of one single minutiae i.e., center minutiae. The authentication of a minutia is assured by the characteristics of its five neighbors. The characteristics are minutiae type, distance to the central minutiae, relative angle calculated by the coordinates of central point, coordinates of neighbor, direction of central point and ridge count. The special matching criteria incorporate fuzzy logic to select final minutiae for matching score calculation.

Anil K. Jain et al. [9] proposed algorithm to compare the latent fingerprint image with that of the stored in the template. From the latent fingerprint minutiae orientation field and quality map are extracted. Both level 1 and 2 features are employed in computing matching scores. Quantitative and qualitative scores are computed at each feature level. Xuzhou Li and Fei Yu [10] proposed fingerprint matching algorithm that uses minutiae centered circular regions. The circular regions constructed around minutiae are regarded as a secondary feature. This feature is tolerant to the linear distortion. Other features like ridge count, distance, relative angle and the minutiae type are also considered to construct the local features. The degree of similarity is obtained based on the local features matching. The minutiae pair that has the higher degree of similarity than the threshold is selected as reference pair minutiae. Jian-De Zheng et al. [11] introduced fingerprint matching based on minutiae. The proposed algorithm uses a method of similar vector triangle. The ridge end points are considered as the reference points. Using the reference points the vector triangles are constructed. The fingerprint matching is performed by comparing the vector triangles. Avinash Pokhriyal and Sushma Lehri [12] proposed an algorithm of fingerprint verification based on wavelets and pseudo Zernike moments. Wavelet was used to denoise and extract ridges. The pseudo Zernike moments was used to extract features which carry the descriptive information about the fingerprint image.

Zhang quinghui and Zhang Xiangfie [13] proposed the algorithm for fingerprint Identification. Fingerprint image pretreatment processes like gamma controller standardization, Directional diagram computation, image filtering, binarization processes and image division were applied for improving the image quality. Xu Cheng and CHENG Xin-Ming [14] proposed the algorithm for fingerprint identification based on wavelet transform and Gabor features. Center point area of the fingerprint image is detected by calculating Poincare index value and region of interest. The image around the centre of fingerprint is decomposed into sub images using wavelet transform to get more features. From the sub images the Gabor wavelet features were extracted to generate feature vector for matching. M Dadgostar et al., [15] presented a fingerprint identification method where features were extracted using Gabor filter and recursive Fischer linear discriminant. The fingerprint image was decomposed into several windows of 32 * 32 pixels and normalized using mean and variance. Using Gabor filter in different orientations features were extracted. To reduce the high dimensional features, recursive Fischer linear discriminant was applied to extract more discriminant features. Classification being achieved using Nearest Cluster Centre classifier with Leave one out method and 3NN classifier. Mohammed Khalil et al., [16] presented statistical analysis of fingerprint images for personal identification. The fingerprint image was enhanced using short time Fourier transform analysis. After detecting the reference point, a sub image of 129 * 129 pixels was extracted by taking the reference point as centre. The statistical technique is used to compute a Gray Level Co-occurrence Matrix (GLCM). The multiple GLCM were computed for different angles and thus four features were extracted for matching. Elsa Timothy Anzaku et al., [17] proposed multifactor authentication using fingerprint and user specific pseudo random number such as secret key. The fingerprint feature extraction was performed using global and local ridge characteristics to generate a fixed length code. Reference core point was detected based on complex filtering. The binding of user specific pseudo random number and fingerprint data was performed using random projection to create feature vector. Euclidean distance was used for matching. Nae myo [18] presented the algorithm of multilayer convex polygon for fingerprint Identification. The fingerprint image was subjected to enhancement and segmentation for noise removal. Using computational geometry algorithms multiple convex layers were created. Among the entire layers smallest polygon was considered. An ellipse which covers the smallest polygon and its all points was considered. The reference polygon and the area ratio have been extracted for matching. Shaharam Mohammadi and Ali Farajzadeh [19] suggested an approach for fingerprint reference point detection using orientation field and curvature measurements. The fingerprint is subjected to region masks. The approximate location of singular points was determined by applying the orientation field and edge detection algorithms. The exact location of the singular point was detected using curvature measurement where the value increases near the singular point and decreases away from it. Using secondary filter and edge detection algorithms reference point has been detected. Bhupesh Gour et al., [20] used ART1 clustering algorithm and Modular neural network for fingerprint recognition. The data base of fingerprint was classified into number of classes by using ART1 clustering algorithm. The monolithic neural network and Back propagation algorithm was used to train the entire fingerprint in the data base. The test fingerprint was compared with the class to which the test fingerprint belongs to. The fingerprint recognition was performed by both monolithic and modular neural network and their performance being compared. Zhang Yuanyuan and Jing Xiaojun [21] presented the algorithm of spectral analysis based fingerprint image enhancement. The characteristics of Gabor filter functions i.e., spectral spatial analysis was performed to estimate the Gabor filter's parameters, filter design and filtering. Then in the frequency domain the fingerprint images were filtered using the Gabor frequency domain filter function. Yanan Meng [22] proposed the algorithm for an improved adaptive pre-processing for fingerprint image.

Hasan Fleyeh et al., [23] presented an algorithm for segmentation of low quality fingerprint images. Noisy, complex and low quality fingerprint images were enhanced using Gaussian filter and histogram equalization. During segmentation process the fingerprint image was divided into sub images of 10 * 10 pixels then local mean, local variance and local coherence were computed. Using these parameters the image is segmented into foreground and background area. Post processing technique was employed to eliminate the presence of isolated blocks. Chi ma et al., [24] proposed a method based on principal curve to extract fingerprint skeleton. The principal graph algorithm was added by the Vertices Merge step, Fitting – Smoothing step, projection step and Vertex optimization step to get the skeleton of the fingerprint. Conti et al., [25] proposed algorithm of finding singularity points for efficient fingerprint classification. The method extracts the singularity points like core, delta and pseudo singularity points using poincare indexes to achieve classification. Hamming distance was used for matching score. The single rotation matching are the two algorithms used for matching. Zhou Weina et al., [26] introduced the algorithm based on wavelet and edge detection for fingerprint verification. The fingerprint image was decomposed to get the norms of high frequency coefficients. The prewitt edge detector on the images gives the edge pixel sets. Matching was performed using threshold values.

## 3. Model

The definitions of performance parameters and the proposed DWTFR model for DWT based Fingerprint Recognition using Non Minutiae Features (DWTFR) is discussed in detail.

3.1 Definitions.

**3.1.1 Correlation:** It is a measure of how correlated a pixel is to its neighbor over the whole image. The range for Gray-Level Co-occurrence Matrix (GLCM) is given by [-1 1]. Correlation is 1 or -1 for a perfectly positively or negatively correlated image. Correlation is NaN (Not-a-Number) for a constant image. It is given by the Equation 1.

$$Correlation = \sum_{i=1}^{N}\sum_{j=1}^{M} \frac{(i-\mu_i)(j-\mu_j)\,P(i,j)}{\sigma_i\,\sigma_j} \quad \ldots\ldots\ldots (1)$$

**3.1.2 Contrast:** It is a measure of the intensity contrast between a pixel and its neighbor over the whole fingerprint image; it is given by Equation 2. The range for Gray-Level Co-occurrence Matrix is given by [0, (size (GLCM,1)-1)^2], Contrast is zero for a constant image.

$$Contrast = \sum_{i=1}^{N}\sum_{j=1}^{N} (|i-j|)^2 P(i,j) \quad \ldots\ldots\ldots\ldots (2)$$

**3.1.3 Energy:** It is the sum of squared elements in the GLCM and given by Equation 3. The range for GLCM is given by [0 1], Energy is 1 for a constant image.

$$Energy = \sum_{i=1}^{N}\sum_{j=1}^{N} P(i,j)^2 \quad \ldots\ldots\ldots\ldots\ldots (3)$$

**3.1.4 Homogeneity:** It is a value that measures the closeness of the distribution of elements in the GLCM to the diagonal and given by the Equation 4. The range for gray-level co-occurrence matrix is [0 1], homogeneity is 1 for a diagonal GLCM.

$$Homogeneity = \sum_{i=1}^{N}\sum_{j=1}^{N} p(i,j)/(1+(i-j)) \quad \ldots\ldots\ldots (4)$$

**3.1.5 False Acceptance Rate (FAR):** It is the probability that an unauthorized person is incorrectly accepted as authorized person using Match Count (MC) and product of Number of Fingers (NF) with Total Images per Finger(IF).

$$FAR\% = \frac{MC}{(NF*IF)} * 100 \quad \text{--------(5)}$$

**3.1.6 False Rejection Rate (FRR):** It is the probability that the system does not detect an authorized person using Miss Match Count (MMC) and Number of Fingers (NF).

$$FRR\% = \frac{MMC}{NF} * 100 \quad \text{--------(6)}$$

**3.1.7 Equal Error Rate (EER):** It is the rate at which both accept and reject rates are equal.

**3.1.8 Total Success Rate (TSR):** It is the rate at which match occurs successfully.

$$TSR\% = \frac{MC}{NF} * 100 \quad \text{------(7)}$$

3.2 Proposed DWTFR Model:

DWT based Fingerprint recognition using Non Minutiae Features is as shown in Fig 1.

3.2.1 Fingerprint image

Fingerprint images are considered from the data base of FVC 2004. The fair and distinct fingerprint image databases DB1, DB2, DB3 and DB4 are created with different scanners and time as shown in the Fig 2. Each data base has 110 fingers with 8 samples per finger leading to 880 fingerprint images. DB3 data base is considered to test our algorithm and it is decomposed into two parts viz., DB3_A and DB3_B having first 100 fingers and last 10 fingers respectively. The source data base consists of DB3_A with first seven samples of every fingerprint constituting 700 samples. The test fingerprint data base consists of DB3_A with the last eighth sample of each fingerprint leads to 100 fingerprint samples. An eight fingerprint samples of second persons in DB3_A database is shown in the Fig 3. Using the source data base and test data base, FRR can be calculated. The DB3_B data base is considered for second test data base having 80 samples which are used to compute FAR.

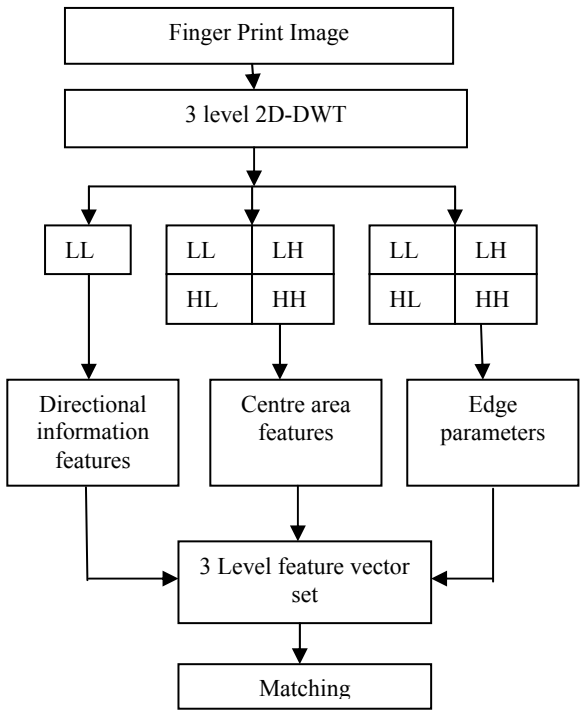

Fig1. Block Diagram DWTFR

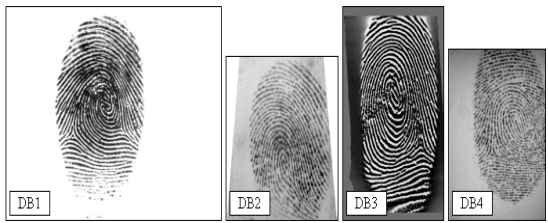

Fig 2. Fingerprint database

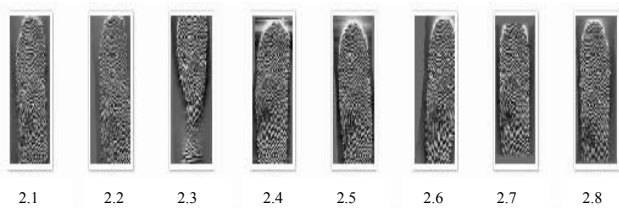

Fig 3. A sample of finger print of DB3_A

### 3.2.2 Discrete Wavelet Transform (DWT)

The fingerprint image is decomposed into multi resolution representation using DWT. The three level Daubechies wavelet is applied and features are extracted from LL, LH, HL and HH sub bands for the verification of fingerprint. LL sub band gives over all information of the original fingerprint image, LH sub band represents vertical information of the fingerprint image, HL gives horizontal characteristics of the fingerprint image and HH gives diagonal details of the fingerprint image.

### 3.2.3 Directional Information Features.

The LL sub band of the DWT is considered for directional information features. The gradient of LL is computed using LH and HL sub bands i.e., the gradient $G_{mn}$ and corresponding angle $\theta_{mn}$ at the position (m, n) is computed using Equations 8 and 9 respectively.

$$G_{mn} = \left(|G_{mn}^x| + |G_{mn}^y|\right) \quad \text{---------(8)}$$
$$\theta_{mn} = \tan^{-1}\left(G_{mn}^x / G_{mn}^y\right) \quad \text{------(9)}$$

The quantities $G_{mn}^x$ and $G_{mn}^y$ represent the components of $G_{mn}$ in horizontal and vertical directions, respectively. The coherence is determined using gradient and angle as given in the Equation. 10 using the window size of (5 x 5).

$$\delta_{mn} = \frac{\sum G_{ij} \cos(\theta_{mn} - \theta_{ij})}{\sum G_{ij}} \quad \text{--------(10)}$$
Where I = 1 to 5
And j = 1 to 5

The dominant local orientation is calculated from the gradient and coherence. The dominant local orientation angle $\theta$ is defined in Equation 11.

$$\theta = \frac{1}{2}\tan^{-1}\frac{\sum_{m=1}^{N}\sum_{n=1}^{N}\delta_{mn}^2 \sin 2\theta_{mn}}{\sum_{m=1}^{N}\sum_{n=1}^{N}\delta_{mn}^2 \cos 2\theta_{mn}} + \frac{\pi}{2}$$

$$\text{----------- (11)}$$

Where N = 8.
Thus, each 8x8 size window represents one directional information. Fig 4 and 5 show the Coherence and Dominant local orientation respectively. From the resultant Dominant local orientation and Coherence matrices Correlation, Contrast, Homogeneity and Energy are computed.

### 3.2.4 Center Area Features

The Four sub bands of wavelet LL, LH, HL and HH are considered for centre area features. The centre point for LL, LH, HL and HH sub bands are fixed by considering the pixel with maximum variance among rows and columns. The mean of the i[th] row $\overline{d\iota}$, the variance of the $i^{th}$ row $s_i^2$, the mean of the $j^{th}$ column $\overline{d_j}$, and the variance of the $j^{th}$ column $s_j^2$, are computed using the statistical Equation 12 through 15. The point at which the highest variance of row as well as the highest variance of column is met is picked up as the centre point.

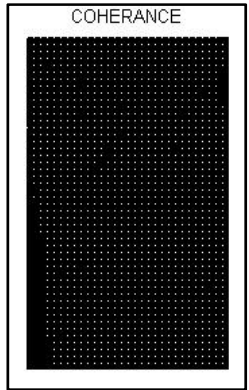

Fig 4 Coherence image

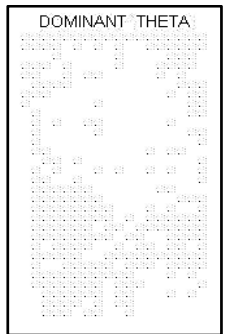

Fig 5 Dominant local orientation image

$$\overline{d_i} = 1/M \sum_{j=1}^{N}(d_{ij})$$

………… (12)

$$S^2_i = 1/M \sum_{j=1}^{N}(d_{ij} - \overline{d_i})^2$$

………… (13)

$$\overline{d_j} = 1/M \sum_{i=1}^{N}(d_{ij})$$

………… (14)

$$S^2_j = 1/M \sum_{i=1}^{N}(d_{ij} - \overline{d_j})^2$$

………… (15)

The 16 * 16 window is considered around the centre point. The Correlation, Contrast, Homogeneity and Energy are determined for 16 * 16 windows around the centre point for all four sub bands to derive the second set of fingerprint features.

### 3.2.5 Canny's Edge Detection

The canny's edge algorithm is applied to all the four LL, LH, HL and HH sub bands. The performance of the Canny's edge algorithm depends heavily on the adjustable parameters, σ (Standard deviation) which is the standard deviation for the Gaussian filter, and the threshold values, T1 and T2. σ also controls the size of the Gaussian filter.

### 3.2.6 Feature Vectors

Directional information features, centre area features and canny's edge parameters are concatenated to constitute feature vector set with one level DWT decomposition. Similarly the feature vector set consisting of Directional features, Centre area features and canny's edge parameters are also obtained for second and third level DWT decomposition. The final feature is the combination of feature vectors of all the three level DWT decomposition.

### 3.2.7 Matching

The final feature vector of test fingerprint is determined and is compared with the feature vector sets of fingerprint database. The Euclidean Distance is computed between fingerprint data base and test fingerprint. The matching and non matching is based on the prefixed Euclidean Distance which constitutes threshold value i.e., if Euclidean Distance is less than the threshold value then it is matched else not matched.

## 4. Algorithms

Problem Definition: The Fingerprint is verified using DWTFR algorithm.

The objectives are

i) The finger print is verified using directional features of wavelet domain and edge parameters.

ii) The value of FAR, FRR and EER be reduced.
iii) The value of TSR be increased

Table 1: Algorithm for DWTFR

> - Input: Finger print Image
> - Output: Verified Finger print image
> i) Read finger print data base of DB3_A
> ii) Three level DW3 decomposition is applied on finger print
> iii) For each level DWT decomposition directional information features, Centre area features and canny edge parameters are obtained to get initial feature vector.
> iv) The final feature vector is the concatenation of three feature vectors of three levels DWT decomposition.
> v) The fingerprint match/non match is based on Euclidean distance and threshold values.

Assumptions: The fingerprint data base DB3_A of FVC 2004 database having finger print size of 300 * 480 with 512 dpi is considered for performance analysis. The algorithm of DWTFR is shown in the Table 1 for fingerprint verification by concatenating directional information features, centre area features and canny's edge parameters.

## 5. Performance Analysis

The DB3_A fingerprint data base is considered for the performance analysis. The performance test parameters such as FAR, FRR and TSR are computed and are given in the Table 2 for different threshold values. It is observed that the FAR decreases and FRR increases as threshold values increases.

The variations of FAR and FRR with different threshold values are shown in the Fig 6. The FAR decreases and FRR increases with threshold values. The EER value is 1.98% at which FRR and FAR become equal at a threshold value of 58.5. The percentage value of FAR, FRR and TSR given in the Table 3 for existing algorithm of *A New Method of Fingerprint Authentication using 2D Wavelets* (AMFAUW) [12] and proposed DWTFR algorithms. The proposed algorithm has improved values for FAR, FRR and TSR compared to the existing algorithms.

Table 3: Comparison of FAR, FRR and TSR for existing and proposed algorithm

| Existing Method AMFAUW | | | Proposed Method DWTFR | | |
|---|---|---|---|---|---|
| %FAR | %FRR | %TSR | %FAR | %FRR | %TSR |
| 5.91 | 6.14 | 94.09 | 0 | 3 | 97 |

Table 2: FAR, FRR and TSR for various threshold values

| Threshold | %FAR | %FRR | %TSR |
|---|---|---|---|
| 25 | 14 | 0 | 100 |
| 30 | 7 | 0 | 100 |
| 35 | 7 | 0 | 100 |
| 40 | 7 | 0 | 100 |
| 45 | 7 | 0 | 100 |
| 50 | 7 | 0 | 100 |
| 55 | 7 | 0 | 100 |
| 57 | 7 | 0 | 100 |
| 59 | 0 | 2 | 98 |
| 60 | 0 | 3 | 97 |
| 61 | 0 | 8 | 92 |
| 62 | 0 | 15 | 85 |

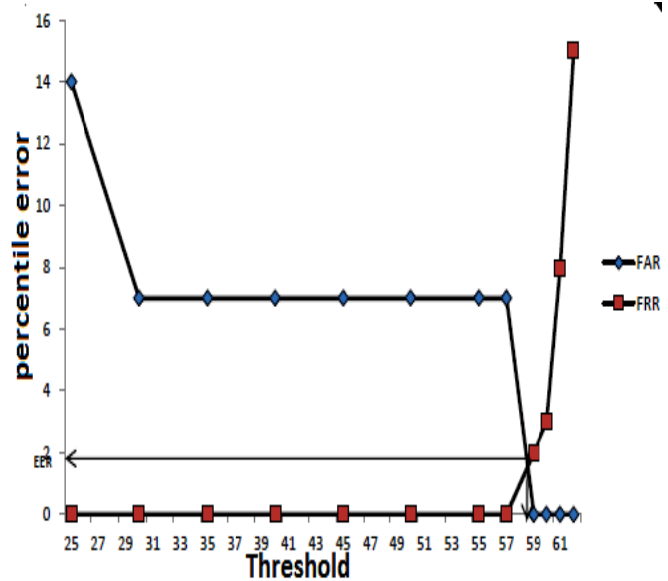

Fig 6. Graph of variations of FAR and FRR.

## 6. Conclusion

Fingerprint image can be acquired by a co-operative and non co-operative person, hence it is easy to create a database for security purpose. In this paper DWT based Fingerprint Recognition using Non Minutiae Features is proposed. The features of fingerprint such as Directional Information, Centre Area and Edge Parameters are extracted from all the three DWT levels by considering four sub bands. The matching between test and data base fingerprint are verified using Euclidean distance. The FAR, FRR and TSR values are better in case of proposed algorithm compared to the existing algorithm. In future the algorithm may be used for spatial domain and other transform domains.

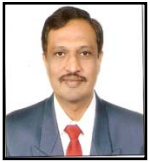
**Shashikumar D R** received BE degree in Electronics & Communication Engineering from Mysore University and ME degree in Electronics from Bangalore University, Bangalore. He is pursuing his Ph.D. in Information and Communication Technology of Fakir Mohan University, Balasore, Orissa under the guidance of Dr. K. B. Raja, Assistant Professor, Department of Electronics and Communication Engineering, University Visvesvaraya College of Engineering, Dr.Sabyasachi Pattanaik Reader & HOD, Department of Information and Communication Technology F M University, Balasore, Orissa, R K Chhotaray, Principal, Seemantha Engineering College, Orissa. He has got 6 publications in national and International Journals and Conferences. He is currently working as Professor, Dept. of Computer Science, Cambridge Institute of Technology, Bangalore. His research interests include Microprocessors, Pattern Recognition, and Biometrics.

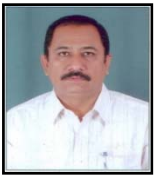
**K B Raja** is an Assistant Professor, Dept. of Electronics and Communication Engineering, University Visvesvaraya college of Engineering, Bangalore University, Bangalore. He obtained his BE and ME in Electronics and Communication Engineering from University Visvesvaraya College of Engineering, Bangalore. He was awarded Ph.D. in Computer Science and Engineering from Bangalore University. He has received best paper presentation awards in many International Conferences. He is involved in guiding several Ph.D scholars in the field of Image processing, VLSI design and Computer Networks. He has over 60 research publications in refereed International Journals and Conference Proceedings. His research interests include Image Processing, Biometrics, VLSI, Signal Processing and Computer Networks.

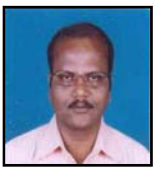
**Dr. Sabyasachi Pattnaik** has done his B.E in Computer Science, and M Tech., from IIT Delhi. He has received his Ph D degree in Computer Science in the year 2003 and now working as Reader in the Department of Information and Communication Technology, in Fakir Mohan University, Vyasavihar, Balasore, Orissa, India. He has got 20 years of teaching and research experience in the field of Neural Networks, Soft Cmputing Techniques. He has got 48 publications in National & International Journals and Conferences. He has published three books in Office Automation, Object Oriented Programming using C++ and Artificial Intelligence. At present he is involved in guiding 8 Ph D scholars in the field of Neural Networks, Cluster Analysis, Bio-informatics, Computer Vision & Stock Market Applications. He has received the best paper award & gold medal from Orissa Engineering Congress in 1992 and Institution of Engineers in 2009.

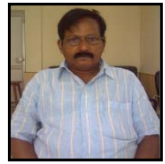
**R K Chhotaray** received B.Sc Engineering in Electrical Engineering and M.Sc Engineering in Electrical Engineering with specialization in Control Systems from Banaras Hindu University, and Ph D in Control Systems from Sambalpur University. He was Professor and Head of Department of Computer Science and Engineering, Regional Engineering College, Rourkela, from which he retired in 2003. Currently he is working as Principal of Seemanta Engineering College, Orissa. He has been associated with many Universities of India in the capacity of Chairman and member of various Boards of Studies, Syllabus Committee, and Regulation Committee. He has about hundred publications in International and National Journals of repute, and has received Best Technical Paper award in many occasions. His special fields of interest include Control of Infinite Dimensional Hereditary Systems, Modeling and Simulation, Theoretical Computer Science, Signal and Image Processing, and Optimization.